\documentclass[letter, 10pt, conference]{ieeeconf}      

\usepackage{amsthm}

\IEEEoverridecommandlockouts                              %
\usepackage{array,multicol,multirow,booktabs,longtable,tabularx}
\usepackage{subfig}
\usepackage{amssymb,mathtools,dsfont,bbm,cuted,stmaryrd}
\usepackage{xspace,algorithmic,algorithm}

\usepackage{tikz}
\usetikzlibrary{math,arrows.meta,spy,automata,intersections,decorations,arrows,decorations.markings,decorations.footprints,fadings,calc,trees,mindmap,shadows,decorations.text,patterns,positioning,shapes,patterns,decorations.pathmorphing,patterns.meta,fit,3d,plotmarks,svg.path}
\usepackage{pgfplots}
\usepgfplotslibrary{patchplots}
\usepackage{tikz-3dplot}
\usepackage{tkz-euclide}

\tikzstyle{valve}=[thick,decoration={markings,  
  mark connection node=dmp,
  mark=at position 0.5 with 
  {
    \node (dmp) [thick,inner sep=0pt,transform shape,rotate=0,minimum width=15pt,minimum height=10pt,draw=none] {};
\draw[thick, fill=white](dmp.north west)--(dmp.south east)--(dmp.north east)--(dmp.south west)--(dmp.north west);
\draw[thick](dmp.center)--++(0,8pt)coordinate(a);
\draw[thick, fill=black] ([shift=(0:6pt)]a) arc (0:180:6pt);
  }
}, decorate]

\tikzstyle{spring}=[thick,decorate,decoration={zigzag,pre length=10,post length=10,segment length=10}]

\tikzstyle{damper}=[thick,decoration={markings,  
  mark connection node=dmp,
  mark=at position 0.5 with 
  {
    \node (dmp) [thick,inner sep=0pt,transform shape,rotate=-90,minimum width=15pt,minimum height=3pt,draw=none] {};
    \draw [thick] ($(dmp.north east)+(2pt,0)$) -- (dmp.south east) -- (dmp.south west) -- ($(dmp.north west)+(2pt,0)$);
    \draw [thick] ($(dmp.north)+(0,-5pt)$) -- ($(dmp.north)+(0,5pt)$);
  }
}, decorate]
\tikzstyle{ground}=[fill,pattern=my grid,draw=none,minimum width=0.75cm,minimum height=0.3cm]

\tikzstyle{box}=[rounded corners=3, minimum height=.75cm, minimum width=1.5cm,draw, semithick, fill=gray!40]
\tikzstyle{gain}=[draw,semithick,isosceles triangle,inner sep=1pt, fill=gray!40]
\tikzstyle{ball}=[fill=gray,draw=black,circle,draw,minimum height=.75cm]
\tikzstyle{arrow}=[ultra thick,-latex]
\tikzstyle{arrowtr}=[,blue, semithick,decoration={markings,mark=at position 0.5 with {\arrow[scale=.7]{triangle 60}}},postaction={decorate}]
\tikzstyle{arrowtrr}=[,blue, semithick,decoration={markings,mark=at position 0.5 with {\arrow[scale=.7]{triangle 60 reversed}}},postaction={decorate}]
\tikzstyle{sum}=[circle,draw,minimum height=8]
\tikzstyle{block}=[rounded corners=3, minimum height=.5cm, minimum width=1cm,draw, semithick,fill=none]

\pgfdeclarepatternformonly[\GridSize]{my grid}{\pgfqpoint{-1pt}{-1pt}}{\pgfqpoint{\GridSize}{\GridSize}}{\pgfqpoint{\GridSize}{\GridSize}}%
{
     \pgfsetlinewidth{0.25pt}
     \pgfpathmoveto{\pgfqpoint{0pt}{0pt}}
     \pgfpathlineto{\pgfqpoint{\GridSize}{\GridSize + 0.1pt}}
     \pgfpathmoveto{\pgfqpoint{0pt}{0pt}}
     \pgfusepath{stroke}
}

\newdimen\GridSize
\tikzset{
    GridSize/.code={\GridSize=#1},
    GridSize=20pt
}

\tikzstyle{normalt}=[font=\small, fill=white,draw=black]
\tikzstyle{linet1}=[color=blue,thick,solid]
\tikzstyle{linet2}=[color=blue,dashed,thick]
\tikzstyle{linet3}=[color=blue,dotted,very thick]
\tikzstyle{linet4}=[color=blue,dashdotted, thick]
\tikzstyle{combt}=[ycomb,color=blue,solid,mark=*,mark options={solid}]

\tikzstyle{axa}=[thin,->]
\tikzstyle{fparrw}=[very thin,-latex',shorten >=.5pt]
\tikzstyle{bfparrw}=[very thin,latex'-latex',shorten <=.5pt,shorten
  >=.5pt]
  
\pgfplotsset{compat=newest}
\pgfplotsset{every tick label/.append style={font=\Large}}
\pgfplotsset{legend style={font=\small}}
\pgfplotsset{
		x label style={at={(axis description cs:0.5,-0.05)},anchor=north, font=\small},
    y label style={at={(axis description cs:-0.05,.5)},rotate=0,anchor=south, font=\small},
  }
\pgfplotsset{every tick label/.append style={font=\normalfont}}
\pgfplotsset{every axis x label/.append style={at={(current axis.south)},below=5mm, font=\large}}
\pgfplotsset{every axis y label/.append style={at={(current axis.west)},yshift=10mm, font=\large}}

\tikzset{
semilog lines/.style={thin, gray},
semilog lines 2/.style={semilog lines,
gray!50 },
semilog half lines/.style={semilog lines 2,
dotted },
semilog label x/.style={semilog lines,
below,font=\small, black},
semilog label y/.style={semilog lines,
right,font=\small,black},
asymp lines/.style={thin, densely dashed,black},
Bode lines/.style={semithick, black}
}
\tikzset{
Nyquist lines/.style={semithick, black},
Nyquist grid/.style={thin,gray},
Nyquist label axes/.style={Nyquist grid,font=\small},
Nyquist label points/.style={font=\small}
}

\tikzset{
        hatch distance/.store in=\hatchdistance,
        hatch distance=10pt,
        hatch thickness/.store in=\hatchthickness,
        hatch thickness=2pt
    }

    \makeatletter
    \pgfdeclarepatternformonly[\hatchdistance,\hatchthickness]{mypattern}
    {\pgfqpoint{0pt}{0pt}}
    {\pgfqpoint{\hatchdistance}{\hatchdistance}}
    {\pgfpoint{\hatchdistance-1pt}{\hatchdistance-1pt}}%
    {
        \pgfsetcolor{green!40}
        \pgfsetlinewidth{\hatchthickness}
        \pgfpathmoveto{\pgfqpoint{0pt}{0pt}}
        \pgfpathlineto{\pgfqpoint{\hatchdistance}{\hatchdistance}}
        \pgfusepath{stroke}
    }
    \pgfdeclarepatternformonly[\hatchdistance,\hatchthickness]{mypattern2}
    {\pgfqpoint{0pt}{0pt}}
    {\pgfqpoint{\hatchdistance}{\hatchdistance}}
    {\pgfpoint{\hatchdistance-1pt}{\hatchdistance-1pt}}%
    {
        \pgfsetcolor{red!40}
        \pgfsetlinewidth{\hatchthickness}
        \pgfpathmoveto{\pgfqpoint{0pt}{0pt}}
        \pgfpathlineto{\pgfqpoint{\hatchdistance}{\hatchdistance}}
        \pgfusepath{stroke}
    }
    \pgfdeclarepatternformonly[\hatchdistance,\hatchthickness]{mypattern3}
    {\pgfqpoint{0pt}{0pt}}
    {\pgfqpoint{\hatchdistance}{\hatchdistance}}
    {\pgfpoint{\hatchdistance-1pt}{\hatchdistance-1pt}}%
    {
        \pgfsetcolor{blue!40}
        \pgfsetlinewidth{\hatchthickness}
        \pgfpathmoveto{\pgfqpoint{0pt}{0pt}}
        \pgfpathlineto{\pgfqpoint{\hatchdistance}{\hatchdistance}}
        \pgfusepath{stroke}
    }

    \makeatletter
\tikzset{hatch distance/.store in=\hatchdistance,hatch distance=5pt,hatch thickness/.store in=\hatchthickness,hatch thickness=5pt}

\pgfdeclarepatternformonly[\hatchdistance,\hatchthickness]{north east hatch}
    {\pgfqpoint{-\hatchthickness}{-\hatchthickness}}
    {\pgfqpoint{\hatchdistance+\hatchthickness}{\hatchdistance+\hatchthickness}}
    {\pgfpoint{\hatchdistance}{\hatchdistance}}%
    {
        \pgfsetcolor{\tikz@pattern@color}
        \pgfsetlinewidth{\hatchthickness}
        \pgfpathmoveto{\pgfqpoint{-\hatchthickness}{-\hatchthickness}}       
        \pgfpathlineto{\pgfqpoint{\hatchdistance+\hatchthickness}{\hatchdistance+\hatchthickness}}
        \pgfusepath{stroke}
    }
\pgfdeclarepatternformonly[\hatchdistance,\hatchthickness]{north west hatch}
    {\pgfqpoint{-\hatchthickness}{-\hatchthickness}}
    {\pgfqpoint{\hatchdistance+\hatchthickness}{\hatchdistance+\hatchthickness}}
    {\pgfpoint{\hatchdistance}{\hatchdistance}}%
    {
        \pgfsetcolor{\tikz@pattern@color}
        \pgfsetlinewidth{\hatchthickness}
        \pgfpathmoveto{\pgfqpoint{\hatchdistance+\hatchthickness}{-\hatchthickness}}
        \pgfpathlineto{\pgfqpoint{-\hatchthickness}{\hatchdistance+\hatchthickness}}
        \pgfusepath{stroke}
    }

\pgfdeclarepattern{
  name=hatch,
  parameters={\hatchsize,\hatchangle,\hatchlinewidth},
  bottom left={\pgfpoint{-.1pt}{-.1pt}},
  top right={\pgfpoint{\hatchsize+.1pt}{\hatchsize+.1pt}},
  tile size={\pgfpoint{\hatchsize}{\hatchsize}},
  tile transformation={\pgftransformrotate{\hatchangle}},
  code={
    \pgfsetlinewidth{\hatchlinewidth}
    \pgfpathmoveto{\pgfpoint{-.1pt}{-.1pt}}
    \pgfpathlineto{\pgfpoint{\hatchsize+.1pt}{\hatchsize+.1pt}}
    \pgfpathmoveto{\pgfpoint{-.1pt}{\hatchsize+.1pt}}
    \pgfpathlineto{\pgfpoint{\hatchsize+.1pt}{-.1pt}}
    \pgfusepath{stroke}
  }
}

\pgfdeclarepattern{
  name=pLines,
  parameters={\hatchsize,\hatchangle,\hatchlinewidth},
  bottom left={\pgfpoint{-.1pt}{-.1pt}},
  top right={\pgfpoint{\hatchsize+.1pt}{\hatchsize+.1pt}},
  tile size={\pgfpoint{\hatchsize}{\hatchsize}},
  tile transformation={\pgftransformrotate{\hatchangle}},
  code={
    \pgfsetlinewidth{\hatchlinewidth}
    \pgfpathmoveto{\pgfpoint{-.1pt}{-.1pt}}
    \pgfpathlineto{\pgfpoint{\hatchsize+.1pt}{\hatchsize+.1pt}}
    \pgfusepath{stroke}
  }
}

\tikzset{
  hatch size/.store in=\hatchsize,
  hatch angle/.store in=\hatchangle,
  hatch line width/.store in=\hatchlinewidth,
  hatch size=5pt,
  hatch angle=0pt,
  hatch line width=.5pt,
}

\tikzset{
  pLines size/.store in=\hatchsize,
  pLines angle/.store in=\hatchangle,
  pLines line width/.store in=\hatchlinewidth,
  pLines size=5pt,
  pLines angle=0pt,
  pLines line width=.5pt,
}

\tikzstyle{stateCB}=[draw=black, circle, fill=cyan!30, text=black, minimum width=3em, line width=1pt]
\tikzstyle{stateP}=[draw,Green, circle, fill=Green!40, text=black, minimum width=3em, line width=1pt]
\tikzstyle{stateV}=[draw,IndianRed, circle, fill=IndianRed!40, text=black, minimum width=3em, line width=1pt]
\tikzstyle{stateD}=[draw,DarkGoldenrod, circle, fill=DarkGoldenrod!40, text=black, minimum width=3em, line width=1pt]
\tikzstyle{stateB}=[draw,LightSlateBlue, circle, fill=LightSlateBlue!40, text=black, minimum width=3em, line width=1pt]

\tikzstyle{fRCB}=[draw=black, fill=red!20, text=black, minimum width=1.5em, minimum height=1.5em,preaction={fill, red!20}, pattern=hatch, pattern color=black, hatch size=7.5pt]
\tikzstyle{fIMU}=[draw=black, text=black, minimum width=1.5em, minimum height=1.5em, preaction={fill, green!10},pattern = dots]
\tikzstyle{fPR-GPS}=[draw,black, text=black, minimum width=1.5em, minimum height=1.5em,preaction={fill, blue!10},pattern=pLines, pattern color=blue, pLines size = 8pt, pLines angle=0]
\tikzstyle{fPR-Galileo}=[draw=black, text=black, minimum width=1.5em, minimum height=1.5em,preaction={fill,lightgray!20}, pattern=north west hatch,hatch distance=7pt,hatch thickness=.5pt,pattern color=black]
\tikzstyle{fDop-GPS}=[draw,black, text=black, minimum width=1.5em, minimum height=1.5em,preaction={fill, orange!10},pattern=pLines, pattern color=orange, pLines size = 8pt, pLines angle=0]
\tikzstyle{fDop-Galileo}=[draw=black, text=black, minimum width=1.5em, minimum height=1.5em,preaction={fill,LightSeaGreen!20}, pattern=north west hatch,hatch distance=7pt,hatch thickness=.5pt,pattern color=LightSeaGreen]
\tikzstyle{fprior}=[draw=black, text=black, minimum width=1.5em, minimum height=1.5em, preaction={fill, orange!30},pattern = bricks]    
\makeatother
\usepackage{ifthen}
\usepackage[svgnames]{xcolor}

\usepackage{placeins}
\usepackage{cuted}

\usepackage{flushend}

\let\labelindent\relax
\usepackage{enumitem}

\newcommand{\be}{\begin{equation}}
\newcommand{\ee}{\end{equation}}

\def\bal#1\eal{\begin{align}#1\end{align}}
\def\baln#1\ealn{\begin{align*}#1\end{align*}}

\newcommand{\ben}{\begin{equation*}}
\newcommand{\een}{\end{equation*}}

\newcommand{\bbm}{\begin{bmatrix}}
\newcommand{\ebm}{\end{bmatrix}}

\newcommand{\bBm}{\begin{Bmatrix}}
\newcommand{\eBm}{\end{Bmatrix}}

\newcommand{\bvm}{\begin{vmatrix}}
\newcommand{\evm}{\end{vmatrix}}

\newcommand{\bVm}{\begin{Vmatrix}}
\newcommand{\eVm}{\end{Vmatrix}}

\newcommand{\bpm}{\begin{pmatrix}}
\newcommand{\epm}{\end{pmatrix}}

\newcommand{\bnm}{\begin{matrix}}
\newcommand{\enm}{\end{matrix}}

\newcommand{\bi}{\begin{itemize}}
\newcommand{\ei}{\end{itemize}}

\newcommand{\bse}{\begin{subequations}}
\newcommand{\ese}{\end{subequations}}

\newenvironment{proof-sketch}{\noindent{ \textit{Sketch of Proof}:}\hspace*{0.5em}}

\theoremstyle{plain}

\newtheorem{prop}{Proposition}

\newtheorem{lem}{Lemma}

\newtheorem{rem}{Remark}

\newcommand{\includeFGO}[2][0.75]{\begin{tikzpicture}[baseline=-2pt]\node[#2, scale=#1] at(0,1pt){};\end{tikzpicture}}
\title{\LARGE \bf Real-time tightly coupled GNSS and IMU integration \\ via Factor Graph Optimization$^{*}$}

\author{Radu-Andrei Cioac\u a$^{1}$, Paul Irofti$^{2,3}$, Cristian Rusu$^{2,3}$,\\ Gianluca Caparra$^{4}$, Andrei-Alexandru Marinache$^{5}$, Florin Stoican$^{1,3, \dagger}$
\thanks{$^{1}$University Politehnica of Bucharest, Romania}%
\thanks{$^{2}$Three Tensors S.R.L., Romania}%
\thanks{$^{3}$University of Bucharest, Romania}%
\thanks{$^{4}$Navigation Systems Definition Section (TEC-SEN), ESA.}%
\thanks{$^{5}$Romanian InSpace Engineering S.R.L. (RISE), Romania.}%
\thanks{* The work is carried out under ESA NAVISP Element 1, devoted to the development of innovative PNT systems, technologies, algorithms and techniques. Three of the authors are supported in part by the project “Romanian Hub for Artificial Intelligence - HRIA”, Smart Growth, Digitization and Financial Instruments Program, 2021-2027, MySMIS no. 351416.}%
\thanks{$\dagger$ Corresponding author: {florin.stoican@}\{upb.ro,three-tensors.com\}.}%
}

\usepackage{graphicx}      

\usepackage{amsfonts,amssymb}
\usepackage{xcolor}

\usepackage[algo2e, linesnumbered, algoruled]{algorithm2e}

\usepackage{hyperref}

\begin{document}

\maketitle

\begin{abstract}
Reliable positioning in dense urban environments remains challenging due to frequent GNSS signal blockage, multipath, and rapidly varying satellite geometry. While factor graph optimization (FGO)–based GNSS–IMU fusion has demonstrated strong robustness and accuracy, most formulations remain offline. In this work, we present a real-time tightly coupled GNSS–IMU FGO method that enables causal state estimation via incremental optimization with fixed-lag marginalization, and we evaluate its performance in a highly urbanized GNSS-degraded environment using the UrbanNav dataset.
\end{abstract}




\section{Introduction}


Radio navigation based on GNSS is the most widely used positioning technology for vehicular and pedestrian applications. However, due to the very low received power of satellite signals, GNSS performance degrades significantly—or even becomes unavailable—in dense urban environments. This degradation is primarily caused by severe multipath interference and frequent non-line-of-sight (NLOS) reception. 
Such conditions substantially limit the reliability of GNSS-based navigation and constrain autonomous operations that usually impose stringent service availability requirements. In contrast, inertial measurement units (IMUs) provide measurements of the platform’s specific force and angular velocity, enabling high-rate estimation of relative motion through inertial navigation. IMU-based navigation is immune to environmental interference and offers accurate short-term positioning, but suffers from bias accumulation over time. Fusing GNSS and IMU data in loosely coupled (LC) or tightly coupled (TC) architectures leverages the complementary characteristics of the two sensors~\cite{groves_principles_2015}.

Most GNSS–IMU fusion systems today rely on variants of the Kalman filter. However, due to the nonlinear nature of the measurement and motion models, single-pass estimators may limit the achievable accuracy. 


Factor Graph Optimization (FGO) provides a probabilistic framework for state estimation in which states and sensor measurements are represented as nodes and factors in a graph structure. The estimation problem is formulated as a nonlinear least-squares optimization yielding a maximum a posteriori (MAP) solution. Due to its flexibility and robustness, FGO is widely used for multi-sensor fusion in challenging environments, particularly in robotic localization and SLAM~\cite{dellaert_factor_2017}, where it is often viewed as an alternative to classical Kalman filter–based methods~\cite{frey_factor_1997, xin_comparative_2023}. A key advantage of FGO lies in its ability to perform multi-pass optimization and temporal smoothing, allowing past states to be refined as new measurements become available.

FGO has been applied to applications involving GNSS and IMU data. Prior work has explored incremental smoothing and switchable constraints to mitigate large residuals caused by multipath effects~\cite{sunderhauf_switchable_2013}, as well as factor-graph formulations based on double-differenced GNSS measurements using a reference station~\cite{wen_towards_2021}. Comparisons between LC and TC FGO architectures have also been reported, demonstrating substantial accuracy gains for TC formulations~\cite{wen_it_2020}. More recently, \cite{ahmadi_adaptive_2025} proposed a TC FGO architecture employing robust loss functions based on Barron’s function, achieving significant improvements in positioning performance. In addition, ~\cite{suzuki_open-source_2025} introduced an open-source FGO library supporting the processing of various raw GNSS measurements.

Despite the demonstrated accuracy benefits of factor-graph smoothing approaches, most existing FGO formulations treat navigation as a predominantly offline or smoothing problem, where the full trajectory is recovered after all measurements are available. As discussed by ~\cite{groves_principles_2015}, smoothing is particularly beneficial in scenarios with poor GNSS availability or low-grade inertial sensors, as future measurements help calibrate inertial errors and reduce drift during GNSS outages. However, this reliance on non-causal information limits applicability in real-time systems, where navigation solutions must be provided with low latency and using only causal measurements. As a result, an open question remains regarding the practical real-time capabilities of FGO-based GNSS–IMU fusion and the trade-offs between smoothing accuracy and operational constraints.

Building upon \cite{cioaca_ecc}, we propose a real-time TC FGO GNSS–IMU navigation method that enables causal state estimation. Real-time operation is achieved via incremental fixed-lag smoothing, relying on iSAM2 \cite{isam2}. The approach integrates raw GNSS data and estimates attitude without relying on additional sensors.

\section{Factor Graphs}

Factor Graph Optimization (FGO) is a probabilistic estimation framework that represents complex inference problems using a bipartite graphical model composed of variable nodes (unknown states to be estimated) and factor nodes (probabilistic constraints derived from external inputs, system dynamics, or prior information)~\cite{frey_factor_1997}. In the navigation context, FGO provides a flexible alternative to recursive filtering approaches. Unlike Kalman filter–based methods, which propagate only the current state and its covariance, FGO optimizes over a window of past states simultaneously. This allows the estimator to re-linearize nonlinear measurement models multiple times and to exploit temporal correlations between states, leading to improved consistency and robustness in highly nonlinear or perturbed environments.

As an illustration, let $\mathcal{X} = \{x_0, x_1, \dots, x_n\}$ denote the set of states over an estimation window, where $x_k$ is the system state at discrete time index~$k$. Each state is connected to measurements $z_k$ through measurement models $h_k(\cdot)$, and to neighboring states through motion models $f_k(\cdot)$. The variable nodes encode the state vectors $x_k$, while the functions $f_k$ and $h_k$ define the factor nodes. Under the common assumption of independent Gaussian noise, the Maximum A Posteriori (MAP) estimate of the state trajectory is obtained by minimizing a nonlinear least-squares problem of the form
\begin{equation}
	\begin{aligned}
    J(\mathcal{X}) = 
    \left\| x_0 - \bar{x}_0 \right\|_{\Sigma_0}^2 
    &+ \sum_{k=1}^n \left\| h_k(x_k) - z_k \right\|_{\Sigma_R^{k}}^2 \\
	&+ \sum_{k=1}^n \left\| f_k(x_{k-1}) - x_k \right\|_{\Sigma_Q^{k}}^2 ,
	\end{aligned}
	\label{eq:FGO}
\end{equation}
where $\bar{x}_0$ is the prior estimate of the initial state, and $\Sigma_0$, $\Sigma_R^{k}$, and $\Sigma_Q^{k}$ are the corresponding covariance matrices. The Mahalanobis norm is defined as $\|r\|_{\Sigma}^2 = r^\top \Sigma^{-1} r$.

The structure of the factor graph allows the cost function~\eqref{eq:FGO} to be decomposed into multiple residual blocks, enabling efficient numerical solutions using iterative methods such as Gauss–Newton or Levenberg–Marquardt. Incremental solvers, such as iSAM2~\cite{isam2}, further exploit this sparsity by selectively re-linearizing and updating only the affected portions of the graph as new measurements arrive. When factor graphs are applied to dynamical systems, the influence of old measurements and states on new ones typically decays over time. This enables the use of marginalization techniques, where past states are removed from the FGO while their information is preserved through a condensed prior. Marginalization effectively bounds both the computational and memory requirements of the optimization problem.

FGO provides a unifying framework for multi-sensor fusion, allowing measurements and motion constraints to be incorporated in an organized and modular manner. Its ability to balance estimation accuracy, robustness, and computational efficiency has led to widespread adoption in robotics and navigation. Moreover, the combination of incremental updates and marginalization makes FGO particularly suitable for real-time estimation problems, where computational efficiency and memory usage are critical.


\section{Tightly Coupled GNSS/IMU via FGO}

Tightly coupled schemes have several significant advantages over their loosely coupled counterparts. In particular, for navigation in challenging environments they enable: (i) maximal fusion of information from a propagation source (e.g. kinematic model or IMU), (ii) maximal GNSS control over constellations, frequencies, models, and general receiver settings, and (iii) inherent outlier damping and increased solution availability. These benefits come at a cost: tighter coupling can introduce convergence issues, demands higher computational power, is more complex to implement, and requires receivers that output raw measurements.

\begin{figure}[!ht]
    \centering
    \includegraphics[width=\columnwidth]{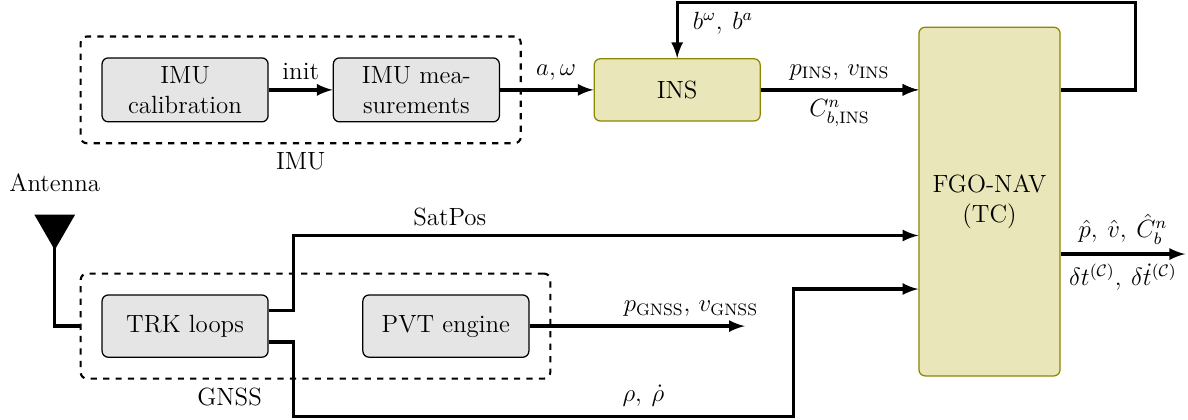}
    \caption{Tightly coupled GNSS/IMU integration scheme.}
    \label{fig:fgo-lc-scheme}
\end{figure}

Hereinafter, we propose an FGO-based tightly coupled scheme, as in Fig.~\ref{fig:fgo-lc-scheme}, which relies on raw measurements from both an IMU and a GNSS receiver. The two information sources are fused in a tightly coupled architecture that combines pre-integrated IMU measurements with observables from all satellites in view. The estimator jointly solves the nonlinear GNSS navigation equations while estimating IMU biases and receiver clock biases within a single optimization problem. This formulation offers key advantages, including robust operation under under-constrained GNSS conditions (e.g. few available satellites or poor geometry). Moreover, the joint formulation naturally fits the FGO framework, enabling batch processing with multi-iteration optimization for improved re-linearization and estimation accuracy.

The resulting factor graph is depicted in Fig.~\ref{fig:fgo_scheme}, and its main components are described in the following sections.

\subsection{State vector}

\begin{figure*}[!htp]  
  \centering
  \includegraphics[width=.875\textwidth]{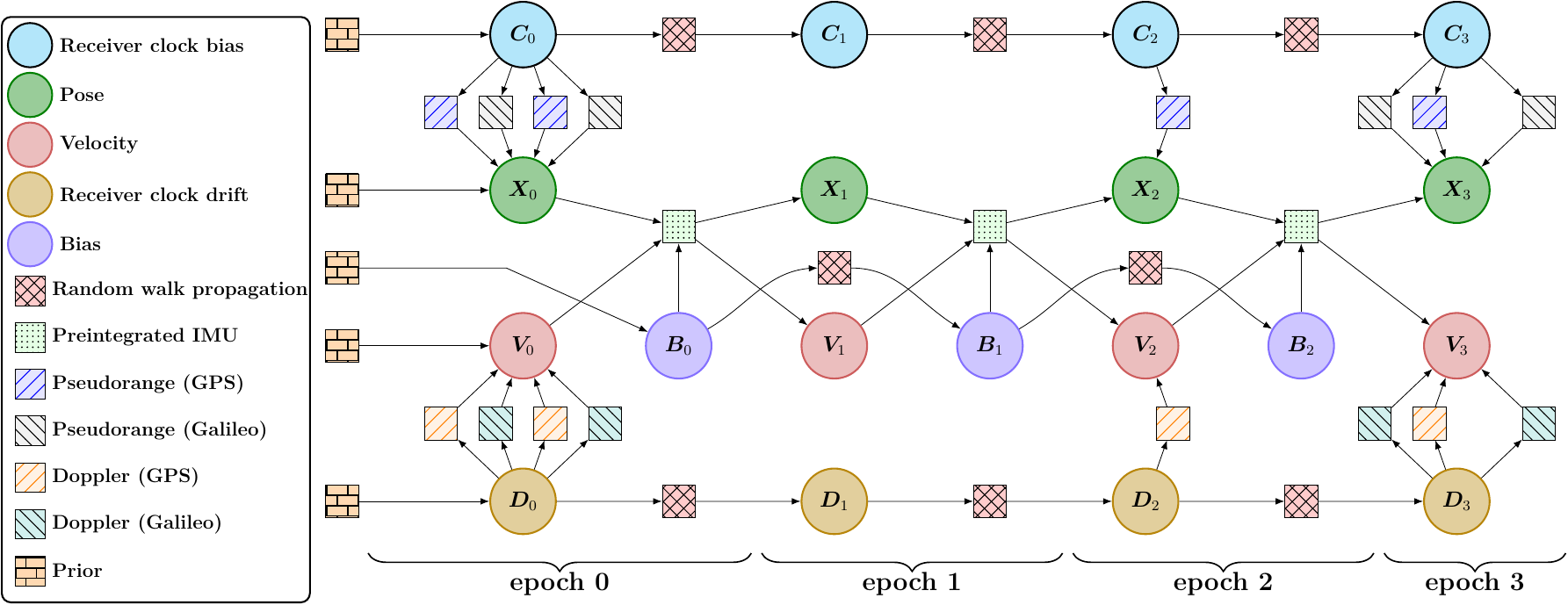}
  \begin{tikzpicture}[transform shape, scale=.575]
  \end{tikzpicture}
  \caption{Tightly coupled GNSS and IMU factor graph architecture.}
  \label{fig:fgo_scheme}AS
\end{figure*}

The state $x_k$ to be estimated is defined as:
\begin{equation}
x_k \triangleq \big( p^{n}_k, \; v^{n}_k, \; C^{n}_{b,k}, \; b^{a}_k, \; b^{\omega}_k, \; \delta t^{(\mathcal C)}_k, \; \delta \dot t^{(\mathcal C)}_k \big),
\end{equation}
and is composed of: $p^n_k\in\mathbb{R}^3$, the position in the navigation frame; $v^n_k\in\mathbb{R}^3$, the velocity in the navigation frame; $C^n_{b,k}\in\mathrm{SO}(3)$, the rotation matrix from body to navigation; $b^a_k\in\mathbb{R}^3$, the accelerometer bias on each axis; $b^{\omega}_k\in\mathbb{R}^3$, the gyroscope bias on each axis; $\delta t^{(\mathcal C)}_k,\ \delta \dot t^{(\mathcal C)}_k \in \mathbb{R}^{n_{\text{const}}}$ are the stacked receiver clock bias and drift (one entry for each constellation used, $c\in\mathcal C$). These appear as variable nodes in Fig.~\ref{fig:fgo_scheme}, as follows: (i) pose (\includeFGO[.3]{stateP}), (ii) velocity (\includeFGO[.3]{stateV}), (iii) IMU bias (\includeFGO[.3]{stateB}), (iv) clock bias (\includeFGO[.3]{stateCB}), and (v) clock drift (\includeFGO[.3]{stateD}).

\subsection{Factors}
Each measurement introduces a factor that constrain the overall optimization problem via its residual and appropriate Mahalanobis norm. 

\smallskip
\noindent {\itshape{Prior factor} (\includeFGO[.65]{fprior}).}
Prior factors are used to anchor the factor graph and define the reference frame by constraining a subset of state variables at the initial epoch:
\begin{equation}
\label{eq:j_prior}
J_{\text{prior}} =
\| x_0 - \bar{x}_0 \|_{\Sigma_0}^2 ,
\end{equation}
where $x_0$ denotes the initial state and $\bar{x}_0$ its prior estimate. 

\smallskip
\noindent {\itshape IMU preintegration factor (\includeFGO[.65]{fIMU}).}
Inertial measurements between consecutive GNSS epochs $k$ and $k{+}1$ are summarized using IMU preintegration. This factor constrains the relative motion between states while accounting for inertial biases:
\begin{equation}
J_{\text{IMU},k} =
\big\|
r_{\text{IMU}}(x_k, x_{k+1})
\big\|_{\Sigma_{\text{IMU}}}^2 ,
\label{eq:j_imu}
\end{equation}
where $r_{\text{IMU}}(\cdot)$ denotes the standard preintegrated IMU residual fully described by~\cite{forster_imu_2015}.

\smallskip
\noindent {\itshape Random walk propagation factor (\includeFGO[.65]{fRCB}).}
Slowly varying states are modeled as discrete-time random walks:
\begin{equation}
J_{\text{RW},k} =
\big\|
s_{k+1} - s_k
\big\|_{\Sigma_s}^2 ,
\label{eq:j_rw}
\end{equation}
where $s_k$ denotes a generic state evolving according to a random walk. We use these factors to propagate IMU biases, receiver clock bias, and receiver clock drift.

\smallskip
\noindent {\itshape GNSS pseudorange factors (\includeFGO[.65]{fPR-GPS}, \includeFGO[.65]{fPR-Galileo}).}
Each GNSS pseudorange observation constrains the receiver position in the Earth-Centered Earth-Fixed (ECEF) frame and the receiver clock bias. Prior to factor construction, raw pseudorange measurements are corrected for all effects that can be compensated, including satellite clock offsets, relativistic effects, Sagnac correction, and signal group delays.

The pseudorange residual for a satellite $m$ is defined as
\begin{equation}
r_{\rho,m} =
\rho_m -
\Big(
\| p^{e}_k - p^{e}_{s,m} \|
+ \delta t^{(\mathcal C)}_k
+ T_m
+ I_m
\Big),
\end{equation}
where $p^{e}_k$ is the receiver position in ECEF, $p^{e}_{s,m}$ is the satellite position, $\delta t^{(\mathcal C)}_k$ is the receiver clock bias associated with the satellite constellation, and $T_m$ and $I_m$ denote tropospheric and ionospheric corrections, respectively.
In the GPS-only case, $\delta t^{(\mathcal C)}_k$ reduces to the GPS receiver clock bias. When multiple constellations are used, the clock state is parameterized by a GPS clock bias and an additional inter-system time offset (e.g., GPS--Galileo time offset), added to the GPS clock bias for non-GPS measurements. The factor cost is
\begin{equation}
J_{\text{PR},m} =
\| r_{\rho,m} \|_{\Sigma_{\text{PR}}}^2 .
\label{eq:j_pr}
\end{equation}

\smallskip
\noindent {\itshape GNSS Doppler factors (\includeFGO[.65]{fDop-GPS}, \includeFGO[.65]{fDop-Galileo}).}
Doppler (pseudorange-rate) observations constrain the receiver velocity and clock drift. The Doppler residual for a satellite $m$ is defined as
\begin{equation}
r_{\dot{\rho},m} =
\dot{\rho}_m -
\Big(
\mathbf{u}_m^\top v^{e}_k
+ \delta \dot t^{(\mathcal C)}_k
\Big),
\end{equation}
where $\dot{\rho}_m$ is the measured pseudorange rate, $\mathbf{u}_m$ is the line-of-sight unit vector from receiver to satellite expressed in ECEF, $v^{e}_k$ is the receiver velocity in ECEF, and $\delta \dot t^{(\mathcal C)}_k$ is the constellation-specific receiver clock drift.
Analogously to the pseudorange case, the clock drift state reduces to a single GPS clock drift in the single-constellation case, and includes an additional inter-system drift term when multiple constellations are used. The associated factor cost is
\begin{equation}
J_{\text{Dopp},m} =
\| r_{\dot{\rho},m} \|_{\Sigma_{\text{Dopp}}}^2 .
\label{eq:j_dopp}
\end{equation}

\subsection{Optimization cost}
Given the active fixed-lag window $\mathcal{W}_k$, the estimation problem, using \eqref{eq:j_prior}--\eqref{eq:j_dopp}, is posed as
\begin{multline}
\label{eq:fgo_cost}
\min_{\mathcal{X}_{\mathcal{W}_k}} \; J(\mathcal{X}_{\mathcal{W}_k})
= {} 
J_{\text{prior}} + \sum_{\ell \in \mathcal{W}_k} J_{\text{IMU},\ell} + \sum_{\ell \in \mathcal{W}_k} J_{\text{RW},\ell} \\[2pt]
 + \sum_{\ell \in \mathcal{W}_k}
    \sum_{m \in \mathcal{S}_\ell}
    J_{\text{PR},(\ell,m)}
 + \sum_{\ell \in \mathcal{W}_k}
    \sum_{m \in \mathcal{S}_\ell}
    J_{\text{Dopp},(\ell,m)},
\end{multline}
where $\mathcal{X}_{\mathcal{W}_k}$ denotes the set of state variables within the window, $\mathcal{S}_\ell$ is the set of satellites tracked at epoch $\ell$. In the incremental fixed-lag smoother, states older than the window are marginalized, and their information is preserved through the resulting prior on the remaining variables.

\subsection{Standard FGO Approach (SFGO)}

The SFGO framework, introduced in a loosely coupled form in~\cite{ahmadi_loosely_sfgo} and later extended to a tightly coupled GNSS–IMU architecture in~\cite{ahmadi_adaptive_2025}, formulates the navigation problem as a nonlinear smoothing problem under a least-squares objective. Prior factors, IMU preintegration factors, and GNSS pseudorange factors are combined in a single factor graph, and the full state trajectory is recovered by minimizing a global cost over all measurements. By exploiting future information, SFGO attains high accuracy and robustness, often outperforming EKF-based approaches. However, the smoothing formulation introduces latency, since solutions are only available after processing a batch of data, making SFGO mainly suitable for post-processing or offline applications where accuracy is favored over real-time operation.

Related FGO-based formulations have also been proposed. In~\cite{wen_towards_2021}, raw GNSS observations (including carrier phase) are used to achieve RTK-level accuracy without an IMU, while~\cite{wen_factor_2021} and ~\cite{wen_it_2020} presents a tightly coupled factor-graph scheme that uses IMU-measured accelerations and an external heading sensor, omitting gyroscope measurements.

\subsection{Proposed Approach: RTFGO Tightly Coupled}

Building upon the SFGO formulation and extending our previously proposed real-time loosely coupled approach (RTFGO-LC)~\cite{cioaca_ecc}, we introduce a real-time, tightly coupled FGO method, denoted RTFGO-TC. The proposed approach adapts the factor-graph optimization framework to support causal, real-time state estimation.

To achieve this, three key modifications are introduced. First, real-time operation is obtained by enforcing causality through the use of the iSAM2 incremental solver within a fixed-lag smoothing framework. By marginalizing out states older than a predefined lag, the estimator outputs navigation solutions using only causal measurements, trading the benefits of full batch smoothing for immediate solution availability while bounding computational complexity and slightly trading off estimation accuracy.

Second, pseudorange-rate measurements are incorporated through dedicated GNSS Doppler factors. This provides additional constraints on receiver velocity and clock drift, leading to improved velocity estimation and, in turn, enhanced attitude and position accuracy.

Third, unlike approaches that rely on externally provided attitude estimates or explicit attitude constraints (e.g., \cite{wen_factor_2021}), the proposed method does not inject attitude information from additional sensors. Instead, attitude and inertial bias states become observable indirectly through their dynamic coupling with velocity and position, as in classical tightly coupled INS/GNSS architectures~\cite{groves_principles_2015}. As a consequence, full observability requires trajectories with sufficient dynamics (e.g., turns or accelerations), which may lead to longer transient convergence in low-dynamic scenarios, but avoids the need for external attitude constraints or additional sensors.

\section{Experimental Results}

We present an experimental evaluation of the proposed method in a real-world urban scenario, focusing on positioning accuracy, service availability, and real-time computational performance. The implementation is developed in Python on top of the GTSAM library~\cite{gtsam} and executed on a MacBook Air equipped with an Apple M3 processor. Raw GNSS processing is performed using functions adapted from the \texttt{rtklib-py} library\footnote{\url{[https://github.com/rtklibexplorer/rtklib-py}}. To facilitate further research, the source code used in this study is publicly available at \url{https://codeberg.org/3T-NAFGO/TightlyCoupledLocalizationFGO}.

\subsection{Data Source}

\begin{figure*}[!ht]
    \subfloat[Ground truth and GNSS (Loops 1 and 2)]{\label{fig:loops_gt}\includegraphics[trim={600 0 600 0}, clip, width=.67\columnwidth]{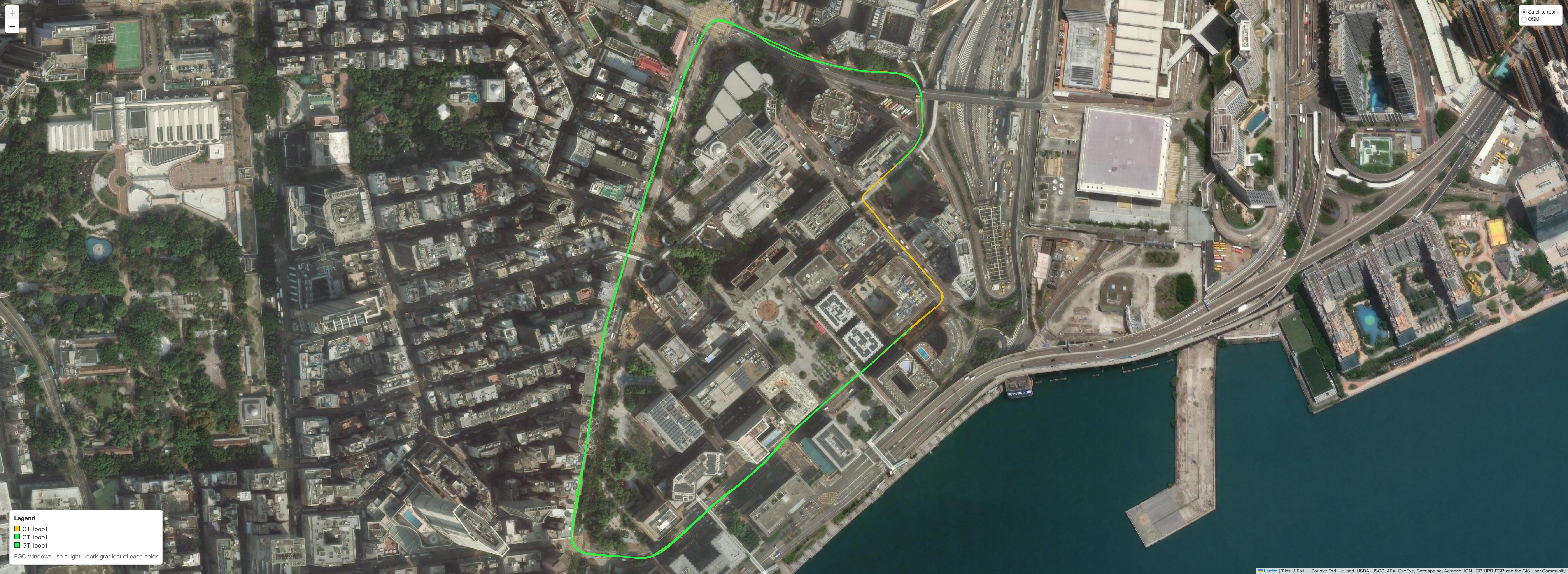}}\hfill
    \subfloat[Estimated trajectories (Loop 1)]{\label{fig:loops_1}\includegraphics[trim={0 0 0 0}, clip, width=.67\columnwidth]{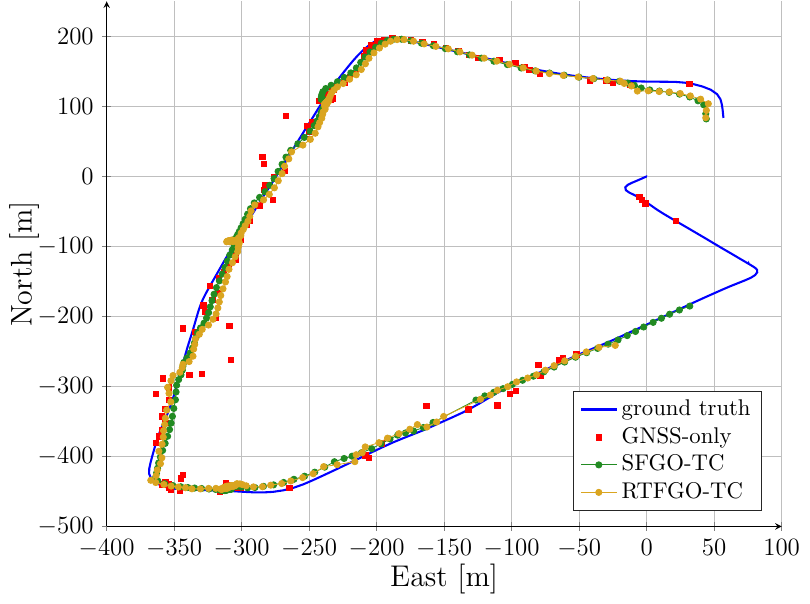}}\hfill
    \subfloat[Estimated trajectories (Loop 2)]{\label{fig:loops_2}\includegraphics[trim={0 0 0 0}, clip, width=.67\columnwidth]{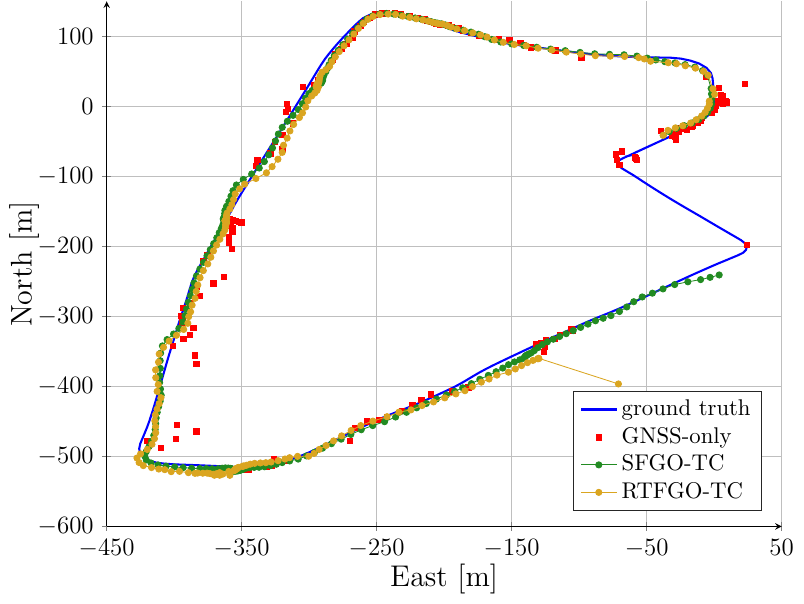}}
    \caption{Comparison of SFGO and RTFGO performance against GNSS-only and ground truth data (Loops 1 and 2)}
\end{figure*}

The experimental evaluation is conducted using the \emph{UrbanNav-HK-MediumUrban-1} dataset from the UrbanNav benchmark suite~\cite{Hsunavi.602}. This dataset was collected in a dense urban environment in Hong Kong and is designed to test GNSS-based positioning algorithms under challenging conditions, including severe multipath, frequent NLOS reception, and partial satellite blockage (see Figure~\ref{fig:loops_gt}). The dataset provides raw GNSS measurements together with high-rate IMU data and a high-accuracy ground-truth trajectory. The vehicle traverses the same path twice, forming two distinct loops (\emph{Loop~1} and \emph{Loop~2}, see Table~\ref{tab:dataset-segments}), enabling repeatability analysis under similar environmental conditions.

\begin{table}[!ht]
\centering
\caption{Dataset loops and GNSS availability.}
\begin{tabular}{lcccc}
\toprule
Segment & Start time (GPST) & Duration [s] & GNSS Availability [\%] \\
\midrule
Loop~1 & 02:33:31 & 336 & 42.0 \\
Loop~2 & 02:39:13 & 422 & 38.9 \\

\bottomrule
\end{tabular}
\label{tab:dataset-segments}
\end{table}

\begin{figure}[!ht]
\centering
\includegraphics[width=\columnwidth]{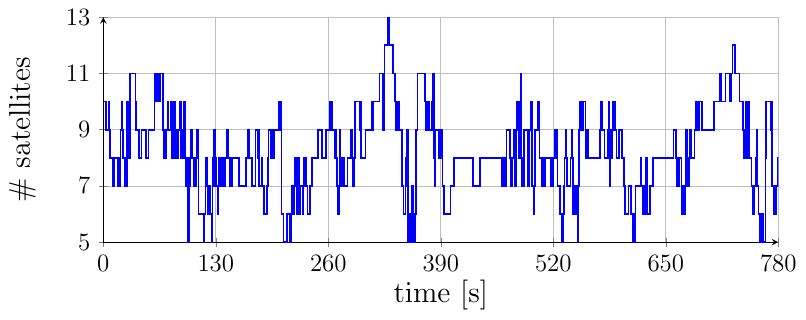}
\caption{Number of tracked satellites (along Loops 1 and 2).}
\label{fig:n_sats}
\end{figure}

A GNSS-only solution was computed using the Single Point Positioning (SPP) mode of RTKLIB~\cite{RTKLIB} to characterize the impact of the urban environment on standalone GNSS performance. The results show a generally low GNSS availability in this configuration (see Fig.~\ref{fig:loops_1} and Fig.~\ref{fig:loops_2}), mainly due to frequent residual thresholding and chi-square outlier rejection performed by RTKLIB. Notably, the number of tracked satellites exhibits strong variability (see Fig.~\ref{fig:n_sats}).

Based on this analysis, a short segment of the trajectory was intentionally excluded from the evaluation (marked in Fig.~\ref{fig:loops_gt} with orange). It corresponds to an extremely dense urban canyon where SPP provides almost no valid position fixes and all tested methods exhibit significantly increased residual errors due to prolonged signal blockage and severe attenuation. Removing this segment prevents the results from being dominated by a scenario in which reliable positioning is not feasible and ensures a fair comparison across approaches.

\subsection{Results}

The proposed approach (RTFGO-TC) is evaluated against the GNSS-only solution computed using RTKLIB, our real-time LC implementation (RTFGO-LC)~\cite{cioaca_ecc}, and the SFGO framework in both loosely coupled (SFGO-LC) and tightly coupled (SFGO-TC) configurations. The SFGO results are obtained in batch mode by disabling real-time constraints. Two-dimensional trajectory results for both loops and all evaluated methods are shown in Fig.~\ref{fig:loops_1} and Fig.~\ref{fig:loops_2}. The metrics reported below are computed using an infinite marginalization window, i.e., no states are marginalized during optimization. 

\begin{figure}[!ht]
    \centering
    \includegraphics[width=.85\columnwidth]{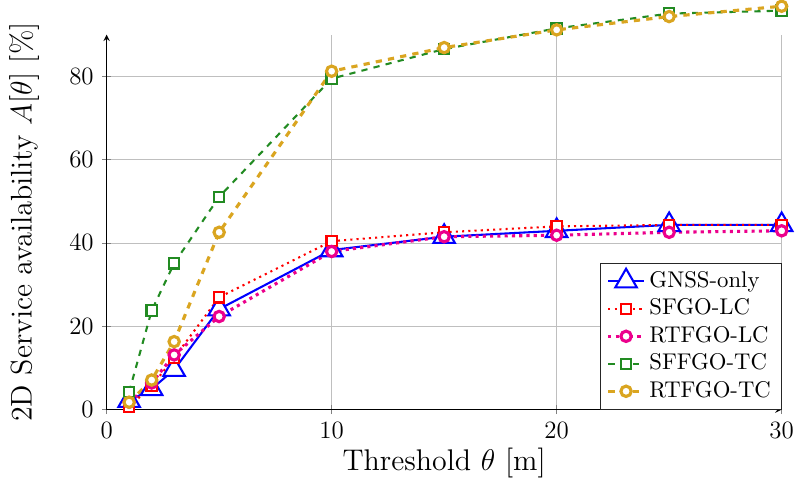}
    \caption{2D Service Availability}
    \label{fig:service_availability}
\end{figure}

\emph{Service availability} measures the fraction of time for which a valid position estimate is provided within a specified error bound. Availability curves are reported for 2D RMSE position errors in Fig.~\ref{fig:service_availability}. They illustrate how different fusion architectures trade positioning accuracy against robustness to GNSS outages. For example, for a 2D RMSE error threshold of $10 \text{m}$, both SFGO-TC and RTFGO-TC methods provide a service availability of around $80\%$, whereas their loosely coupled and GNSS-only counterparts reach only about $40\%$.

Noteworthy, RTFGO-TC avoids hard measurement rejection at the receiver level and remains operational during periods of degraded satellite visibility, substantially increasing the time for which a valid navigation solution is available.

2D and 3D RMSE positioning accuracy results are reported in Table~\ref{tab:accuracy_loops_2d3d_compact}. Accuracy is compared only at epochs where both the GNSS-only and loosely coupled solutions provide valid estimates. This restriction is necessary to ensure a fair comparison, since the tightly coupled formulations yield a continuous solution over the entire trajectory, including periods where the other methods are unavailable.

\begin{table}[!ht]
\centering
\caption{RMSE accuracy (2D/3D) for Loop~1 and Loop~2}
\begin{tabular}{l r r}
\toprule
Method &
\multicolumn{1}{c}{\begin{tabular}{@{}c@{}}Loop~1\\RMSE 2D/3D [m]\end{tabular}} &
\multicolumn{1}{c}{\begin{tabular}{@{}c@{}}Loop~2\\RMSE 2D/3D [m]\end{tabular}} \\
\midrule
GNSS-only (RTKLIB) & 8.87 / 30.54 & 8.25 / 12.23 \\
RTFGO-LC & 10.92 / 34.25 & 8.24 / 10.75 \\
RTFGO-TC & 6.44 / 29.60 & 7.16 / 22.88 \\
SFGO-LC & 5.87 / 27.78 & 6.44 / 8.12 \\
SFGO-TC & 4.79 / 21.79 & 4.72 / 9.73 \\
\bottomrule
\end{tabular}
\label{tab:accuracy_loops_2d3d_compact}
\end{table}

The proposed RTFGO-TC formulation consistently improves 2D positioning accuracy with respect to both the loosely coupled and GNSS-only solutions. This behavior highlights the benefit of tight GNSS–IMU integration in constraining horizontal motion, particularly in urban environments where satellite visibility is intermittent. In contrast, the 3D RMSE metric shows degradation in some cases, most notably for Loop~2, where the error nearly doubles with respect to the loosely coupled formulation. This increase is primarily driven by larger errors in the vertical component. In dense urban scenarios, GNSS vertical observability is significantly weaker than horizontal observability due to unfavorable satellite geometry. In the tightly coupled formulation, the absence of explicit vertical constraints (e.g., height priors or map-based constraints) allows vertical errors to propagate through the inertial integration, resulting in increased altitude drift during periods of GNSS degradation. While this effect degrades the 3D RMSE, it has limited influence on horizontal positioning accuracy, the primary performance metric for ground-vehicle and pedestrian navigation.

\begin{figure}[t!]
    \centering
    \includegraphics[width=.9\columnwidth]{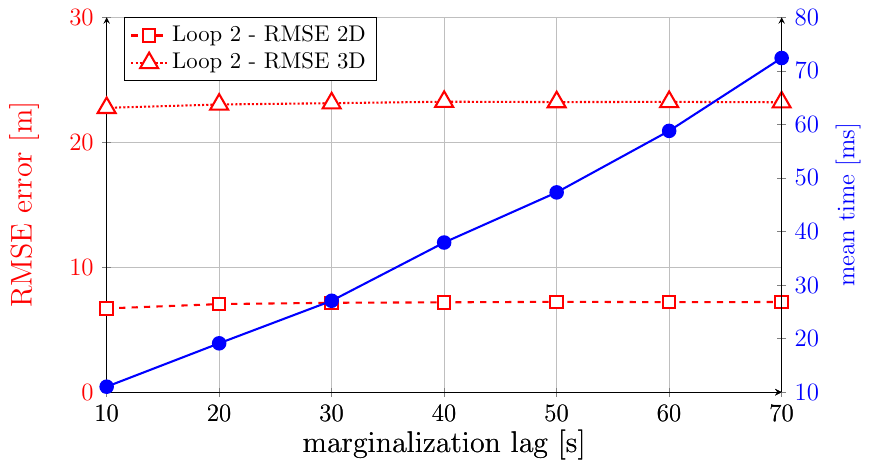}
    \caption{Effect of marginalization lag on positioning accuracy and mean optimization time for RTFGO-TC on Loop~2.}
    \label{fig:marginalization_error_mean_latency}
\end{figure}

The impact of the fixed marginalization window length on estimation accuracy and computational cost is analyzed for Loop~2 and the results are illustrated in Fig.~\ref{fig:marginalization_error_mean_latency}.

Increasing the marginalization lag is expected to improve estimation accuracy by including a longer temporal context. However, the results show that the impact on both 2D and 3D positioning accuracy remains limited, with only small variations observed as the window size increases. This behavior can be attributed to the inclusion of older measurements of reduced quality in larger windows, which may partially offset the benefits of extended temporal smoothing. 

In contrast, the computational cost exhibits a clear increase with the marginalization lag. For a window size of 60 samples, corresponding to one minute of past states retained in the graph, the mean optimization time reaches $\sim\mkern-4mu 59\,\mathrm{ms}$. This increase is significant when compared to the loosely coupled RTFGO-LC implementation, which requires $\sim\mkern-4mu 6\,\mathrm{ms}$ for the same window size~\cite{cioaca_ecc}. These results highlight the inherent trade-off between computational efficiency and estimator consistency in tightly coupled factor-graph formulations. 

\section{Conclusions}

This paper presents RTFGO-TC, a real-time tightly coupled GNSS--IMU factor graph framework that enables real-time estimation while retaining the modeling flexibility of factor graph optimization. By leveraging incremental iSAM2 optimization with fixed-lag marginalization, the proposed approach achieves bounded computational complexity and low latency, making tightly coupled FGO suitable for real-time operation. Raw GNSS pseudorange and Doppler measurements are directly fused with IMU preintegration without relying on external attitude sources, allowing attitude and inertial biases to become observable through system dynamics.

Experimental evaluation on the \textit{UrbanNav} dataset, representative of dense urban GNSS-challenged environments, demonstrates that RTFGO-TC significantly improves service availability and horizontal positioning accuracy compared to GNSS-only and loosely coupled integrations.

Future work will focus on extending the proposed framework to carrier-phase--based positioning methods, including PPP and RTK, by explicitly estimating carrier-phase ambiguities within the factor graph. Existing research~\cite{xin_comparative_2023,wang_factor_2023} suggests promising results in those areas. In addition, we plan to incorporate vehicle- and pedestrian-specific motion constraints, such as non-holonomic constraints, zero-velocity updates, and lever-arm modeling, to further improve robustness and consistency in urban scenarios.

This work has been submitted to the IEEE for possible publication. Copyright may be transferred without notice, after which this version may no longer be accessible

\end{document}